\def\year{2019}\relax
\documentclass[letterpaper]{article} %DO NOT CHANGE THIS
\usepackage{aaai19}  %Required
\usepackage{times}  %Required
\usepackage{helvet}  %Required
\usepackage{courier}  %Required
\usepackage{url}  %Required
\usepackage{graphicx}  %Required

\usepackage{latexsym}
\usepackage{amsmath}
\usepackage{amssymb}
\usepackage{xcolor}
\usepackage{enumerate}
\usepackage{comment}
\usepackage{subfigure}
\usepackage{multirow}
\usepackage{bm}
\usepackage{url}

\begin{document}
% \bstctlcite{BSTcontrol}
\title{View Inter-Prediction GAN: Unsupervised Representation Learning for 3D Shapes by Learning Global Shape Memories to Support Local View Predictions}
\author{Zhizhong Han\textsuperscript{1,2},
Mingyang Shang\textsuperscript{1},
Yu-Shen Liu\textsuperscript{1}\thanks{Corresponding author: Yu-Shen Liu},
Matthias Zwicker\textsuperscript{2}\\
\textsuperscript{1}{School of Software, Tsinghua University, Beijing, China}\\
\textsuperscript{2}{Department of Computer Science, University of Maryland, College Park, USA}\\
h312h@umd.edu,
smy16@mails.tsinghua.edu.cn,
liuyushen@tsinghua.edu.cn,
zwicker@cs.umd.edu
}
%\author{AAAI Press\\
%Association for the Advancement of Artificial Intelligence\\
%2275 East Bayshore Road, Suite 160\\
%Palo Alto, California 94303\\
%}
\maketitle
\begin{abstract}
\begin{quote}
In this paper we present a novel unsupervised representation learning approach for 3D shapes, which is an important research challenge as it avoids the manual effort required for collecting supervised data. Our method trains an RNN-based neural network architecture to solve multiple view inter-prediction tasks for each shape. Given several nearby views of a shape, we define view inter-prediction as the task of predicting the center view between the input views, and reconstructing the input views in a low-level feature space. The key idea of our approach is to implement the shape representation  as a shape-specific global memory that is shared between all local view inter-predictions for each shape. Intuitively, this memory enables the system to aggregate information that is useful to better solve the view inter-prediction tasks for each shape, and to leverage the memory as a view-independent shape representation. Our approach obtains the best results using a combination of $L_2$ and adversarial losses for the view inter-prediction task. We show that VIP-GAN outperforms state-of-the-art methods in unsupervised 3D feature learning on three large scale 3D shape benchmarks.
\end{quote}
\end{abstract}

\section{Introduction}
Feature learning for 3D shapes is crucial for 3D shape analysis, including classification~\cite{Sharma16,WuNIPS2016,Zhizhong2016b,YaoqingCVPR2018,PanosCVPR2018ICML,HanTIP18}, retrieval~\cite{Sharma16,WuNIPS2016,Zhizhong2016b,YaoqingCVPR2018,PanosCVPR2018ICML,HanTIP18}, correspondence~\cite{Zhizhong2016b,HanTIP18} and segmentation~\cite{cvprpoint2017,nipspoint17}. In recent years, supervised 3D feature learning has produced remarkable results under large scale 3D benchmarks by training deep neural networks with supervised information~\cite{cvprpoint2017,nipspoint17}, such as class labels and point correspondences. However, obtaining supervised information requires intense manual labeling effort.
%, what is worse, it cannot be blended from different benchmarks to facilitate better 3D feature learning.
Therefore, unsupervised 3D feature learning with deep neural networks is an important research challenge.

Several studies have addressed this challenge~\cite{Sharma16,WuNIPS2016,Zhizhong2016b,Girdhar16,RezendeEMBJH16,YaoqingCVPR2018,PanosCVPR2018ICML,HanTIP18} by training deep learning models using ``supervised'' information mined from the unsupervised scenario. This mining procedure is usually implemented using different prediction strategies, such as the prediction of a shape from itself by minimizing reconstruction error or embedded energy, the prediction of a shape from its context given by views or local shape features, or the prediction of a shape from views and itself together.
These methods use multiple views to provide a holistic context of 3D shapes, and they make a single global shape prediction based on all views.
%Among these methods, views have been regarded as the context of 3D shapes, where multiple views are used to make a global prediction one single time.

%In contrast, we propose a method that learns to make multiple local view predictions from neighboring views, and aggregates the obtained knowledge from these predictions. One of our main contributions is to show that this leads to a highly discriminative global shape feature with state-of-the-art shape classification and retrieval performance.

In contrast, our approach called \textit{View Inter-Prediction GAN} (VIP-GAN) learns to make multiple local view inter-predictions among neighboring views. The view inter-prediction task is designed to mimic human perception of view-dependent patterns. That is, based on changes between neighbor views, humans can easily imagine the center view between, while the neighbor views can also be reversely imagined based on the center. As a key idea, our network architecture implements the shape representation as a shape-specific global memory whose contents are learned to support all local view inter-prediction tasks for each shape. Intuitively, the memory aggregates information over all view inter-prediction tasks, which leads to a view-independent shape representation. Our experimental results indicate that the obtained representation is highly discriminative and outperforms competing techniques on several standard shape classification benchmarks.

%Intuitively, this strategy captures more and finer-grained discriminative information than a single global shape prediction from multiple views.

%However, this strategy omits more and finer discriminative ``supervised'' information that lies among the multiple views, which could not learn competitive global features as the strategy by the prediction of shape from itself. Therefore, how to predict views locally from their neighboring views and aggregate the knowledge derived from multiple local predictions makes the challenge even harder to overcome.

%However, the involved context is always globally employed as a group, which omits more and finer discriminative ``supervised'' information that could be locally mined in the context, i.e., the prediction of a part of context from another part of context. Therefore, how to perform prediction inside the context and aggregate the knowledge derived from each prediction makes the challenge even harder to overcome.

More specifically, VIP-GAN considers multiple views taken around a 3D shape in sequence as the context of the 3D shape, and it separates each view sequence into several overlapping sections of equal length. It then learns to predict the center view from its neighbors in each section, and the neighbors from the center. Crucially, VIP-GAN includes a memory shared by all view predictions of each shape. We show that the system uses this memory to improve its view prediction performance, in effect by learning a view independent shape representation. VIP-GAN employs an RNN-based generator with an encoder-decoder structure to implement the view inter-prediction strategy in different spaces. The encoder RNN captures the content information and spatial relationship of the neighbors to predict the center in 2D view space, while the decoder RNN predicts the neighbors in a low-level feature space according to the center predicted by the encoder. To further improve the prediction of the center, we train the generator jointly with a discriminator in an adversarial way.
In summary, our significant contributions are as follows:

\begin{enumerate}[i)]
\item We propose VIP-GAN as a novel deep learning model to perform unsupervised 3D global feature learning through view inter-prediction with adversarial training, which leads to state-of-the-art performance in shape classification and retrieval. %enables multiple views to be learned with adversarial strategy as other 3D raw representations in an unsupervised way.
\item VIP-GAN makes it possible to mine fine-grained ``supervised'' information within the multi-view context of 3D shapes by imitating human perception of view-dependent patterns, which facilitates effective unsupervised 3D global feature learning.
\item We introduce a novel implicit aggregation technique for 3D global feature learning based on RNN, which enables VIP-GAN to aggregate knowledge learned from each view prediction across a view sequence effectively.% than the widely used explicit aggregation, such as pooling. MZ: removed this, because we don't directly compare to pooling
\end{enumerate}

\section{Related work}
\noindent\textbf{Supervised 3D feature learning. }Recently, supervised 3D feature learning is an attractive topic. With class labels, various deep learning models have been proposed to learn 3D features from different 3D raw representations, such as voxels~\cite{Wu2015}, meshes~\cite{HanTIP18}, points clouds~\cite{cvprpoint2017,nipspoint17} and views~\cite{tmmbs2016,Bshi2015,Sfikas17,eccvSinha2017,su15mvcnn,JohnsLD16,AsakoCVPR2018}, which aims to capture the mapping between 3D raw representations and class labels. The mapping is captured by spotting the distribution patterns among voxels~\cite{Wu2015}, points in cloud~\cite{cvprpoint2017,nipspoint17}, vertices on mesh~\cite{HanTIP18}, or view features taken from different shapes~\cite{tmmbs2016,Bshi2015,Sfikas17,eccvSinha2017,su15mvcnn,JohnsLD16,AsakoCVPR2018}. Among these methods, multi-view based 3D feature learning methods perform the best, where pooling is widely used for view aggregation.

\noindent\textbf{Unsupervised 3D feature learning. }Although unsupervised 3D feature learning methods~\cite{Sharma16,WuNIPS2016,Zhizhong2016b,Girdhar16,RezendeEMBJH16,YaoqingCVPR2018,PanosCVPR2018ICML,HanTIP18} are not always with high performance as supervised ones, their promising advantage of learning without labels still draws a lot of attention. To mine ``supervised'' information from unsupervised scenario, unsupervised feature learning methods usually train deep learning models by different prediction strategies, such as the prediction of a shape from itself by minimizing reconstruction error~\cite{Sharma16,WuNIPS2016,YaoqingCVPR2018,PanosCVPR2018ICML} or embedded energy~\cite{Zhizhong2016b}, the prediction of a shape from context~\cite{HanTIP18}, or the prediction of a shape from context and itself together~\cite{Girdhar16,RezendeEMBJH16}. These methods employ different kinds of 3D raw representations, such as voxels~\cite{Sharma16,WuNIPS2016,Girdhar16,RezendeEMBJH16}, meshes~\cite{Zhizhong2016b,HanTIP18} or point clouds~\cite{YaoqingCVPR2018,PanosCVPR2018ICML}, and accordingly, different kinds of context, such as spatial context of virtual words~\cite{HanTIP18} or views~\cite{Girdhar16,RezendeEMBJH16}, are employed. With the ideas of auto-encoder~\cite{Sharma16,Girdhar16,RezendeEMBJH16,YaoqingCVPR2018,PanosCVPR2018ICML}, classification~\cite{HanTIP18} or generative adversarial training~\cite{WuNIPS2016,PanosCVPR2018ICML}, these methods effectively learn discriminative 3D features. Different from these methods, VIP-GAN tries to learn 3D features by performing view inter-prediction to mine fine-grained ``supervised'' information within the multi-view context of 3D shapes, where context formed by multiple views is first explored for 3D global feature learning with adversarial training.

\noindent\textbf{View synthesis and unsupervised video feature learning. }View synthesis aims to generate novel views according to existing views. Deep learning based view synthesis has been drawing more and more research interests~\cite{TatarchenkoDB16,WilliamICLR17}. First tries teach deep learning models to predict novel views according to input views and transformation parameters~\cite{TatarchenkoDB16}. To generate views with more detail (i.e. texture) and less geometric distortions, external image sets or geometric constraints are further employed.

Similarly, to predict the future frames in a video, the information of multiple past frames is aggregated by RNN~\cite{WilliamICLR17}. However, these methods mainly focus on the quality of generated views rather than the discriminability of learned features, where we find the view quality is not a sufficient condition for the feature discriminability in our experiments. In addition, the knowledge learned in each prediction cannot be aggregated by these methods to represent the global features. Therefore, these methods cannot be directly used for unsupervised 3D feature learning from view inter-prediction, which highlights our novelty by differentiating VIP-GAN apart from them.

%VIP-GAN is also different from unsupervised video feature learning studies. This is because there is no firm starting position in a view sequence as in a frame sequence because of 3D shape rotation. This requires VIP-GAN to learn the same feature of a 3D shape, no matter which view of the 3D shape is the first. However, unsupervised video feature learning is sensitive to the first frame of a video at test stage.

VIP-GAN is also different from unsupervised video feature learning studies. Sequential views of 3D shapes are different from video frames because there is no firm starting position in view sequences. Each view could be the first view because of 3D shape rotation. This requires VIP-GAN to be invariant to the initial view position, that is, no matter which view of a 3D shape is the first, the learned feature of the shape should be the same. This is the main characteristic that makes VIP-GAN different from unsupervised video feature learning (At test stage, sensitive to the first frame of a video). Similarly, unsupervised image feature learning cannot aggregate multiple views and employ multiple view consistency as VIP-GAN.

\section{VIP-GAN}
\label{sec:vip}
\begin{figure}[hb]
  \centering
  % the following command controls the width of the embedded PS file
  % (relative to the width of the current column)
  %\includegraphics[width=.95\linewidth, bb=39 696 126 756]{figures/definition3.eps}
   \includegraphics[width=\linewidth]{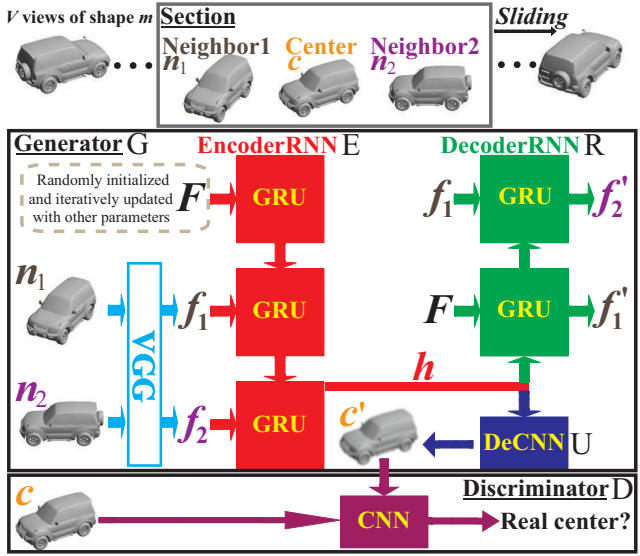}
  % replacing the above command with the one below will explicitly set
  % the bounding box of the PS figure to the rectangle (xl,yl),(xh,yh).
  % It will also prevent LaTeX from reading the PS file to determine
  % the bounding box (i.e., it will speed up the compilation process)
  % \includegraphics[width=.95\linewidth, bb=39 696 126 756]{sampleFig}
  %
  %
\caption{\label{fig:framework} VIP-GAN is composed of generator $\rm G$ and discriminator $\rm D$. The global feature $\bm{F}$ is learned in $\rm G$ by view inter-prediction through encoder $\rm E$, decoder $\rm R$ and deconvolutional net $\rm U$.}
\end{figure}

\noindent\textbf{Overview. }The framework of VIP-GAN is illustrated in Fig.~\ref{fig:framework}. Using multiple local view inter-predictions, VIP-GAN aims to learn a global  representation or feature $\bm{F}$ of a 3D shape $m$ from $V$ views $v_i$
%$v_i$, where $v_i$ are
sequentially taken around $m$, where $i\in[1,V]$. Note that $\bm{F}$ is learned for each shape as an $F$ dimensional vector, effectively serving as a view-independent memory that is used in all local view inter-predictions for the shape. Hence $\bm{F}$ implicitly aggregates the knowledge learned from all sections $\bm{s}_i$ across the $V$ views. Learning $\bm{F}$ is performed via gradient descent together with the other parameters in VIP-GAN, where $\bm{F}$ is randomly initialized.
%and $i\in[1,V]$.
We split the set
%$\mathbf{V}$ formed by the $V$
of views into $V$ sections of equal length, where a section $\bm{s}_i$ is centered at each view $v_i$.
%$v_i$.
We denote the center view of the section as $c$, and its $N$ neighbors as $n_j$, where $j\in[1,N]$ ($N=2$ in Fig.~\ref{fig:framework}). In each section $\bm{s}_i$, VIP-GAN first predicts the center $c$ in 2D space from the neighbors $n_j$. Conversely, it also predicts $n_j$ in feature space from the predicted center $c'$.

VIP-GAN consists of two main components, the generator $\rm G$ and discriminator $\rm D$. The goal of the generator is to predict the center view in each section from its neighbors in image space, and the neighbors from the center in feature space. $\rm G$ consists of a VGG19 network, an encoder RNN $\rm E$ (in red), a decoder RNN $\rm R$ (in green) and a deconvolutional network $\rm U$ (in blue), where $\rm E$ and $\rm R$ are implemented by Gated Recurrent Units (GRUs). In addition, the discriminator $\rm D$ (in purple) is a convolutional network to distinguish whether a center view is real or not. $\rm G$ and $\rm D$ are jointly trained in an adversarial manner.
%to make $\rm G$ generate a more real predicted center $c'$ from $n_j$ in $\bm{s}_i$.

\noindent\textbf{Generator $\rm G$. }In each section $\bm{s}_i$ of shape $m$, the first task of generator $\rm G$ is to collect a feature vector $\bm{h}_i$ that will be used to generate the predicted center view $c'$. For this purpose, the generator encodes the content within the neighbor views $n_j$ and the spatial relationship among them. We extract the content of each $n_j$ as a $4096$ dimensional feature vector $\bm{f}_j$ by the last fully connected layer of a VGG19 network, where the resolution of input $n_j$ is $224\times 224$. We further encode the $\bm{f}_j$ with their spatial relationship using an encoder RNN $\rm E$. We provide the global feature $\bm{F}$ of shape $m$, our learning target, at the first step of the encoder $\rm E$ serving as a knowledge container or memory that keeps incorporating the knowledge derived from each view prediction. %in $\bm{s}_i$ of $m$.
Different from pooling, which is widely used as an explicit view aggregation, this implicit aggregation enables VIP-GAN to learn from more fine-grained information,
%that is inside the context of $m$,
such as the spatial relationship among the neighbors $n_j$ in each section $\bm{s}_i$, and the connection between knowledge derived from different sections $\bm{s}_i$ across $V$ views of $m$. Finally, at the last step of the encoder $\rm E$ for each section $\bm{s}_i$ we obtain a 4096 dimensional feature $\bm{h}_i$ as the hidden state, which we subsequently use to generate the predicted center $c'$ using a deconvolutional network $\rm U$.

By reshaping the 4096 dimensional $\bm{h}_i$ into 256 feature maps of size $4\times 4$, the deconvolutional network $\rm U$ starts generating the predicted center $c'$ with a resolution of $64\times 64$ through four deconvolutional layers. The deconvolutional layers employ 256, 128, 64, and 3 kernels, respectively, and each kernel has size $3\times 3$ and a stride of 2. In each deconvolutional layer, we use a leaky ReLu with a leaky gradient of 0.2.
%is used as non-linear function.
We utilize the $L$-2 loss between the predicted center view ${\rm U}(\bm{s}_i)=c'$ and the ground truth center $c$ to measure the center prediction performance of $\rm G$,
%in the prediction of the center $c$ from the neighbors $n_j$ in $\bm{s}_i$,
%as defined as $L_{\rm U}$ in Eq.~\ref{eq:deconvolution}, where the output ${\rm U}(\bm{s}_i)$ of $\rm U$ is $c'$.
denoted as loss $L_{\rm U}$,
\begin{equation}
\label{eq:deconvolution}
%L_{\rm U}=mean(\|{\rm U}(\bm{s}_i)-c\|^2_2).
%L_{\rm U}=mean(({\rm U}(\bm{s}_i)-c)^2).
L_{\rm U}=\|{\rm U}(\bm{s}_i)-c\|^2_2.
\end{equation}

The second task of generator $\rm G$ is to reversely predict the neighbors $n_j$ from the predicted center $c'$ in each section $\bm{s}_i$. Different from the center view prediction task,
%of prediction of $c$ from $n_j$ in the 2D space,
%this task is conducted in the high-level feature space.
we evaluate the prediction in feature space here.
The two prediction tasks in different spaces enable VIP-GAN to more fully understand the 3D shape $m$. To predict both the content information within each $n_j$ and the spatial relationship among $n_j$ from the predicted center $c'$, we employ a decoder RNN $\rm R$ with $\bm{h}_i$ as initialized hidden state that predicts the features $\bm{f}_j'$ of each neighbor view $n_j$ step by step.
%The feature $\bm{h}_i$ encoded from the high-level feature $\bm{f}_j$ of $n_j$ by encoder $\rm E$ is further decoded to generate the predicted feature $\bm{f}_j'$ of each $n_j$ step by step.
Similar to the encoder $\rm E$, we provide the global feature $\bm{F}$ at the first step of $\rm R$, which is regarded as a reference for the following neighbor feature predictions. Then, $\bm{f}_j'$ is produced at the $j$-th step of $\rm R$ using the feature $\bm{f}_{j-1}$ of its previous counterpart as input. We predict the features $\bm{f}_j'$ in the same order as we provide the corresponding $\bm{f}_j$ to the encoder $\rm E$. We measure the neighbor prediction performance of $\rm G$ using $L$-2 loss in feature space,
\begin{equation}
\label{eq:decoder}
%L_{\rm R}=\frac{1}{N}\sum\nolimits_{j=1}^{N}mean(({\rm R}(\bm{s}_i)_j-\bm{f}_j)^2),
%L_{\rm R}=\frac{1}{N}\sum\nolimits_{j=1}^{N}mean(\|{\rm R}(\bm{s}_i)_j-\bm{f}_j\|^2_2),
L_{\rm R}=\frac{1}{N}\sum\nolimits_{j=1}^{N}\|{\rm R}(\bm{s}_i)_j-\bm{f}_j\|^2_2,
\end{equation}
where ${\rm R}(\bm{s}_i)_j=\bm{f}_j'$ is the output at the $j$-th step of $\rm R$.
%For decoder $\rm R$, the MSE of each dimension between the predicted feature $\bm{f}_j'$ and the feature $\bm{f}_j$ inputted in $\rm E$ is utilized to measure the performance of $\rm G$ in the prediction of the neighbors $n_j$ from the center $c$ in $\bm{s}_i$, as defined as $L_{\rm R}$ in Eq.~\ref{eq:decoder}, where the output ${\rm R}(\bm{s}_i)_j$ at the $j$-th step of $\rm R$ is $\bm{f}_j'$.
In summary, the loss of $\rm G$ is formed by the loss $L_{\rm U}$ of $\rm U$ and the loss $L_{\rm R}$ of $\rm R$.

\begin{table*}[tb]
  \caption{The effects of balance weights $\alpha$ and $\beta$ on the performance of VIP-GAN under ModelNet10.}
  \label{table:balance}
  \centering
  \begin{tabular}{c|ccccccccc}%llllllllll
    \hline
     ($\alpha$,$\beta$) &(1,0.05)&(3,0.05)&(5,0.05)&(3,0.1)&(3,0.01)&(3,0)&(0,0.01)&(0,0)&(0,0)C\\
    \hline
     Instance ACC & 92.73 & \textbf{94.05} & 93.50 & 92.84 & 91.19 & 92.51 & 83.37 & 84.80 & 75.77
    \\
     Class ACC & 92.23 & \textbf{93.71} & 93.01 & 92.50 & 90.62 & 92.08 & 82.05 & 83.96 & 74.78
     \\
    \hline
  \end{tabular}
\end{table*}

\begin{table*}
  \caption{The effects of parameters on VIP-GAN under ModelNet10 in terms of accuracy.}
  \label{table:parameters}
  \centering
  \begin{tabular}{c|ccccccccccc}%lllllllllllll
    \hline
      Parameters&$\rm R$ &$\rm D$&$F(1024)$&$F(2048)$&$F(4096)$&$N(2)$&$N(6)$&$V(6)$& $V(3)$ &cGan&BiDir\\
    \hline
     Instance ACC&90.53&47.80& 92.29 & 92.51 & \textbf{94.05} & 93.17 & 93.50 & 92.62 & 92.51 & 89.10 & 93.83
     \\
     Class ACC& 89.88&44.49& 91.73 & 92.03 & \textbf{93.71} & 92.91 & 93.08 & 92.32 & 92.22 &88.34 & 93.45
    \\
    \hline
  \end{tabular}
\end{table*}

\noindent\textbf{Discriminator $\rm D$. }
%In theory, generator $\rm G$ is capable of mimicing human perception of view varying pattern by view inter-prediction.
%The reason we further employ discriminator $\rm D$ which is jointly trained with generator $\rm G$ in an adversarial way is that,
In preliminary experiments, we found that the quality of predicted center views $c'$ is not a sufficient condition to obtain a highly discriminative global feature $\bm{F}$. For example, a complex and powerful deconvolutional network could generate $c'$ with higher quality than our simple one introduced before, but we found that the learned feature $\bm{F}$ is much less discriminative.
%than the feature learned by our simple one.
This phenomenon is caused by the large capacity of the more complex deconvolutional network to generate high quality view $c'$ from any feature $\bm{h}_i$. However, this may decrease the discriminability of the learned feature $\bm{F}$. What we really want to achieve is that the quality of predicted views $c'$ is mainly due to the discriminability of the learned feature $\bm{F}$, rather than the powerful learning ability of the deconvolutional network.

To resolve this issue, we employ discriminator $\rm D$ with adversarial training to facilitate our simple deconvolutional network $\rm U$.
% MZ: this is repetitive with what was just described above.
%, which enables VIP-GAN to efficiently learn highly discriminative global features $\bm{F}$.
Specifically, $\rm D$ is a CNN with five layers, including four convolutional layers and a one dimensional fully connected layer, where the resolution of input views is $64\times 64$. Each convolutional layer contains 64, 128, 256, 512 kernels respectively, and each kernel has size $5\times 5$ and a stride of 2, where we employ a leaky ReLu with a leaky gradient of 0.2. In the last layer of $\rm D$, a sigmoid function provides the probability that the input is a real center view. Finally, the loss of $\rm D$ is the cross entropy of the probability produced from each $\bm{s}_i$, as defined as $L_{\rm D}$ in Eq.~\ref{eq:discriminator}, where ${\rm D}({\rm U}(\bm{s}_i))$ is the probability that $\rm D$ thinks the predicted center $c'$ from $\bm{s}_i$ by $\rm U$ is real,

\begin{equation}
\label{eq:discriminator}
L_{\rm D}=\log{\rm D}(c)+\log(1-{\rm D}({\rm U}(\bm{s}_i))).
\end{equation}

\noindent\textbf{Adversarial training. }Adversarial training is based on Generative Adversarial Networks (GAN)~\cite{NIPS2014_5423}. The predicted center $c'$ from the generator $\rm G$ is passed to the discriminator $\rm D$ with the real center $c$, where $\rm D$ tries to learn how to distinguish whether a center is real or not. With adversarial training, the discriminator $\rm D$ is trained to maximize the probability when the center is real while minimizing it when the center is generated by generator $\rm G$, as defined in Eq.~\ref{eq:discriminator}. In contrast, the generator $\rm G$ has to be trained to fool the discriminator $\rm D$. Therefore, in $\bm{s}_i$, the loss $L_{{\rm D}2{\rm U}}$ for $\rm G$ from $\rm D$ is defined to make the predicted center ${\rm U}(\bm{s}_i)$ generated by $U$ more real,

\begin{equation}
\label{eq:adversarial}
L_{{\rm D}2{\rm U}}=\log(1-{\rm D}({\rm U}(\bm{s}_i))).
\end{equation}

Finally, we define the loss function of VIP-GAN by combining the aforementioned losses as in Eq.~\ref{eq:VIP}, where the weights $\alpha$ and $\beta$ are used to control the balance among them,
\begin{equation}
\label{eq:VIP}
L=L_{\rm U}+\alpha L_{\rm R}+\beta L_{{\rm D}2{\rm U}}.
\end{equation}
%
%Note that the weight in front of $\rm U$ is kept to be 1, since the contribution of these three losses can be well balanced by merely adjusting $\alpha$ and $\beta$.
Note that simultaneously with the other network parameters, we also optimize the learning target $\bm{F}$ by minimizing $L$ using a standard gradient descent approach by iteratively updating $\bm{F}$ by $\bm{F}\gets \bm{F}-\varepsilon\times \partial L / \partial \bm{F}$, where $\varepsilon$ is the learning rate.

\noindent\textbf{Modes for testing. }Typically, there are two modes of unsupervised learning of features $\bm{F}$ of 3D shapes for testing, which we call the known-test mode and the unknown-test mode. In known-test mode, the test shapes are given with the training shapes at the same time, such that the features of test shapes can be learned with the features of training shapes together. In unknown-test mode, VIP-GAN is first pre-trained under training shapes. At test time, we then iteratively learn the features of test shapes by minimizing Eq.~\ref{eq:VIP} with fixed pre-trained parameters of $\rm U$, $\rm R$ and $\rm D$.

\section{Experimental results and analysis}
\label{section:results_and_analysis}
In this section, the performance of VIP-GAN is evaluated and analyzed. First we discuss the setup of parameters involved in VIP-GAN. These parameters are tuned to demonstrate how they affect the discriminability of learned features in shape classification under ModelNet10~\cite{Wu2015}. Then, VIP-GAN is compared with state-of-the-art methods in shape classification and retrieval under ModelNet10~\cite{Wu2015}, ModelNet40~\cite{Wu2015} and ShapeNet55~\cite{3dor20171050}. All classification is conducted by a linear SVM (with default parameters in scikit-learn toolkit) under the global features learned by VIP-GAN.

\noindent\textbf{Parameter setup. }The balance weights $\alpha$ and $\beta$ are important for the performance of VIP-GAN. In this experiment, we explore the effects of $\alpha$ and $\beta$ on the performance of VIP-GAN under ModelNet10 in terms of average instance accuracy and average class accuracy, as shown in Table~\ref{table:balance}. Initially, the dimension $F$ of global feature $\bm{F}$ is 4096, the center $c$ gets $N=4$ neighbors, and the $V=12$ views of all 3D shapes under ModelNet10 are employed to train VIP-GAN in known-test mode. $\alpha$ and $\beta$ are set to 1 and 0.05, respectively, since they make the initial values of loss $L_{\rm U}$, $L_{\rm R}$ and $L_{{\rm D}2{\rm U}}$ comparable to each other, where a normal distribution with mean of 0 and standard deviation of 0.02 is used to initialize the parameters involved in VIP-GAN.

\begin{figure}[!]
  \centering
  % the following command controls the width of the embedded PS file
  % (relative to the width of the current column)
  %\includegraphics[width=.95\linewidth, bb=39 696 126 756]{figures/definition3.eps}
   \includegraphics[width=\linewidth]{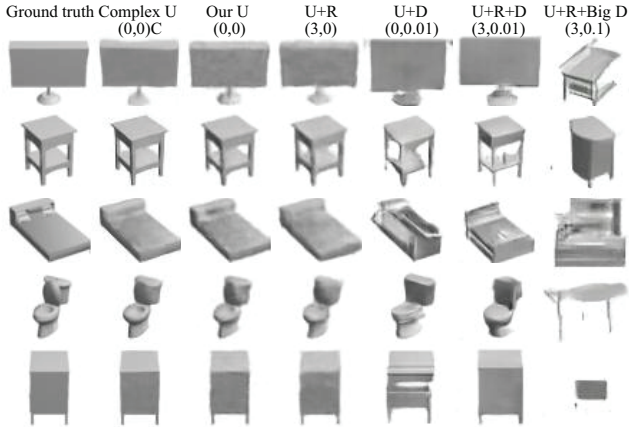}
  % replacing the above command with the one below will explicitly set
  % the bounding box of the PS figure to the rectangle (xl,yl),(xh,yh).
  % It will also prevent LaTeX from reading the PS file to determine
  % the bounding box (i.e., it will speed up the compilation process)
  % \includegraphics[width=.95\linewidth, bb=39 696 126 756]{sampleFig}
  %
  %
\caption{\label{fig:softmax} The predicted centers generated with different pairs of balance parameters $(\alpha,\beta)$.}
\end{figure}

First, the effect of $\alpha$ is explored by incrementally increasing $\alpha$ from 1 to 3 and 5. With $\alpha=3$, best performance of VIP-GAN is achieved up to $94.05\%$, and the results with $\alpha=3$ are better than the results with $\alpha=5$. Then, the effect of $\beta$ is explored based on $\alpha=3$ by increasing $\beta$ to 0.1 and decreasing $\beta$ to 0.01. These degenerated results show that the adversarial loss should not be over- or under-weighted. Subsequently, we highlight the contribution of discriminator $\rm D$ and decoder $\rm R$ to deconvolutional network $\rm U$ by incrementally setting $\alpha$ and $\beta$ to 0. By setting $\beta$ to 0, the results with ``(3,0)'' are better than the results with ``(3,0.01)'', but worse than the results with ``(3,0.1)''. This phenomenon implies that the under-weighted GAN loss is not helpful to increase the discriminability of learned features. We observe a similar phenomenon by comparing between ``(0,0.01)'' and ``(0,0)''. The comparison between ``(3,0)'' and ``(0,0)'' shows that the decoder $\rm R$ significantly increases the discriminability of learned features. In summary, these results show that the decoder $\rm R$ and the discriminator $\rm D$ can both improve the performance of VIP-GAN. However, $\rm R$ contributes more than $\rm D$ to $\rm U$, and $\alpha$ is less sensitive than $\beta$.

Furthermore, as mentioned before, the quality of predicted center $c'$ is not a sufficient condition to obtain a highly discriminative global feature $\bm{F}$. By replacing our simple $\rm U$ with a more complex one employed in~\cite{NIPS2016_6158}, the quality of predicted centers becomes higher, as shown in the comparison between ``(0,0)'' and ``(0,0)C'' in Fig.~\ref{fig:softmax}. On the other hand, the discriminability of the learned global feature $\bm{F}$ dramatically decreases, as illustrated by the comparison between ``(0,0)'' and ``(0,0)C'' in Table~\ref{table:balance}. The reason for this is that the more complex deconvolutional network in~\cite{NIPS2016_6158} is too deep to facilitate effective error back propagation to train a highly discriminative global feature. To keep the network in~\cite{NIPS2016_6158} unchanged, the predicted views are generated in the resolution of $256\times 256$ rather than $64\times 64$, where the $224\times 224$ ground truth views are padded with pixel values of 255 to enable the computation of loss $L_{\rm U}$. Finally, we also highlight the importance of $\rm R$ and $\rm D$ by merely using $L_{\rm R}$ or $L_{{\rm D}2{\rm U}}$ to train, as shown by ``$\rm R$'' and ``$\rm D$'' in Table~\ref{table:parameters}. Compared with the importance of $\rm U$ as ``(0,0)'' in Table~\ref{table:balance}, $\rm R$ plays the most important role in VIP-GAN.

The predicted centers $c'$ generated by different $\alpha$ and $\beta$ are demonstrated in Fig.~\ref{fig:softmax}, where the tags marking each column are consistent with the parameters in Table.~\ref{table:balance}. According to the ground truth, the complex deconvolutional network (``(0,0)C'') generates centers with higher quality than our simple ones (``(0,0)''). The comparison between ``(0,0)'' and ``(3,0)'' shows that the decoder $\rm R$ slightly degenerates the quality of predicted centers. In addition, the adversarial loss weighted by small $\beta$ can make the predicted centers sharper, but also produce distortions, as illustrated by the comparison between ``(0,0)'' and ``(0,0.01)'', and the comparison between ``(3,0)'' and ``(3,0.01)''. The adversarial loss weighted by big $\beta$ will make the loss $L_{\rm U}$ subtle with big distortions, as shown by ``(3,0.1)''.

\begin{table}
  \caption{The comparison of classification accuracy under ModelNet10 and ModelNet40.}
  \label{table:comparison}
  \centering
  \begin{tabular}{cccc}
    \hline
    Methods & Supervised  & MN40 & MN10 \\
    \hline
    MVCNN & Yes & 90.10 & - \\
    MVCNN-Multi & Yes & 91.40 & -\\
    ORION & Yes & - & 93.80 \\
    3DDescriptorNet & Yes & - & 92.40 \\
    Pairwise & Yes & 90.70 & 92.80\\
    GIFT & Yes & 89.50 & 91.50 \\
    PANORAMA & Yes & 90.70 & 91.12 \\
    VoxNet & Yes & - & 92.00\\
    VRN& Yes & 91.33 &93.80 \\
    RotationNet& Yes & 90.65 & 93.84 \\
    PointNet++ & Yes & 91.90 & -   \\
    \hline
    T-L & No & 74.40 & -\\
    LFD& No & 75.47 & 79.90 \\
    Vconv-DAE & No & 75.50 & 80.50 \\
    3DGAN& No & 83.30& 91.00 \\
    LGAN & No & 85.70 & 95.30 \\
    LGAN(MN40) & No & 87.27 & 92.18 \\
    FNet & No& 88.40 & 94.40\\
    FNet(MN40) & No& 84.36 & 91.85\\
    \hline
    Our & No & \textbf{91.98} & \textbf{94.05} \\
    Our1(SN55) & No & 90.19 & 92.18 \\
    Our2(+SN55) & No & 91.25 & 92.84 \\
    \hline
  \end{tabular}
\end{table}

%\begin{table*}
%  \caption{The comparison of classification accuracy under ModelNet10 and ModelNet40.}
%  \label{table:comparison}
%  \centering
%  \begin{tabular}{ccc|ccc}
%    \hline
%    \multicolumn{3}{c|}{ModelNet40} & \multicolumn{3}{|c}{ModelNet10}                   \\
%    \hline
%    Methods & Supervised  & Accuracy & Methods & Supervised  & Accuracy \\
%    \hline
%    MVCNN & Yes & 90.10 & ORION & Yes & 93.80 \\
%    MVCNN-Multi & Yes & 91.40 & 3DDescriptorNet & Yes & 92.40 \\
%    Pairwise & Yes & 90.70 & Pairwise & Yes & 92.80\\
%    GIFT & Yes & 89.50 & GIFT & Yes & 91.50 \\
%    PointNet++ & Yes & 91.90 & VoxNet & Yes & 92.00 \\
%    VRN & Yes & 91.33 &
%    VRN& Yes & 93.80 \\
%    RotationNet & Yes & 90.65 & RotationNet & Yes & 93.84 \\
%    PANORAMA & Yes & 90.70 & PANORAMA & Yes & 91.12 \\
%    \hline
%    T-L Network & No & 74.40 &LFD & No &  79.90   \\
%    Vconv-DAE & No & 75.50& Vconv-DAE & No & 80.50 \\
%    3DGAN & No & 83.30& 3DGAN & No & 91.00 \\
%    LGAN & No & 85.70 &LGAN & No & 95.30 \\
%    LGAN(MN40) & No & 87.27 &LGAN(MN10) & No & 92.18 \\
%    FNet & No& 88.40 &FNet & No & 94.40\\
%    FNet(MN40) & No& 84.36 &FNet(MN10) & No & 91.85\\
%    \hline
%    Our & No & \textbf{91.98} & Our & No & \textbf{94.05} \\
%    Our1(SN55) & No & 90.19 & Our1(SN55) & No & 92.18 \\
%    Our2(+SN55) & No & 91.25 & Our2(+SN55) & No & 92.84 \\
%    \hline
%  \end{tabular}
%\end{table*}

The effects of $F$, $N$ and $V$ are further explored in Table~\ref{table:parameters}. By gradually decreasing $F$ from 4096 to 2048 and 1024, the results are degenerated from $94.05\%$ to $92.51\%$ and $92.29\%$. To conduct this experiment with the rest of VIP-GAN unchanged, one more 4096 dimensional fully connected layer is employed before $\bm{F}$ is inputted in $\rm G$. Then, the number $N$ of neighbors in each section $\bm{s}_i$ is explored by respectively decreasing $N$ to 2 and increasing $N$ to 6, based on the $N=4$ structure with our best results. Although these results are degenerated from our best results, they are still good. The degeneration is caused by that less neighbors could not provide enough discriminative information to learn while more neighbors would bring redundant discriminative information. Following this, we decrease $V$ to 6 and 3 gradually, the results are also decreased due to the less information for learning, where $N$ is adjusted to 2 when $V$ is set to 3. Subsequently, we employ conditional GAN to replace the GAN structure in VIP-GAN, where the ground truth neighbors are regarded as the conditions of the center. The high-level features $\bm{f}_j$ of neighbors are concatenated with the extracted feature of the center after the last convolutional layer in discriminator $\rm D$, which is further followed by an extra convolutional layer and the one dimensional fully connected layer. Although the results dramatically decreased as shown by ``cGan'', it is still better than merely using $\rm U$ as listed ``(0,0)'' in Table~\ref{table:balance}. These results imply that GAN is better than conditional GAN for 3D global feature learning in VIP-GAN, while both the adversarial loss of GAN and conditional GAN are helpful to improve the discriminability of learned features. Moreover, we also try to train VIP-GAN by bidirectional view sequences, since human can perform the view inter-prediction from either left to right or right to left in a view sequence, as shown by the results listed as ``BiDir''. However, no further improvement is obtained from the doubled training samples.

%In this structure, generator $\rm G$ tries to fool discriminator $\rm D$ whether the center is truly surrounded by the neighbors.

\begin{figure}[hb]
  \centering
  % the following command controls the width of the embedded PS file
  % (relative to the width of the current column)
  %\includegraphics[width=.95\linewidth, bb=39 696 126 756]{figures/definition3.eps}
   \includegraphics[width=\linewidth]{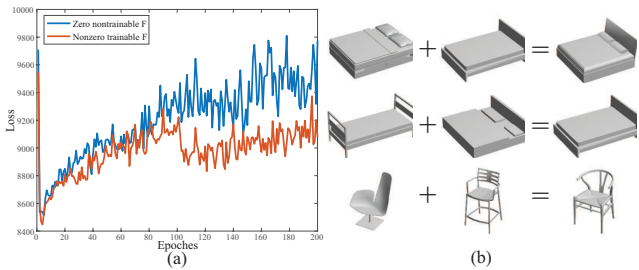}
  % replacing the above command with the one below will explicitly set
  % the bounding box of the PS figure to the rectangle (xl,yl),(xh,yh).
  % It will also prevent LaTeX from reading the PS file to determine
  % the bounding box (i.e., it will speed up the compilation process)
  % \includegraphics[width=.95\linewidth, bb=39 696 126 756]{sampleFig}
  %
  %
\caption{\label{fig:zeros} (a)The effectiveness of our novel implicit view aggregation is shown by the comparison between the loss with nonzero trainable $\bm{F}$ and the loss with zero non-trainable $\bm{F}$. (b)The learned global features are visualized by feature manipulation in the embedding space.}
\end{figure}

\noindent\textbf{Classification. }We compare VIP-GAN with the state-of-the-art methods in classification under ModelNet40 and ModelNet10. The parameters under ModelNet40 are the same ones with our best results under ModelNet10 in Table~\ref{table:parameters}. The compared methods include MVCNN~\cite{su15mvcnn}, ORION~\cite{SZB17a}, 3DDescriptorNet~\cite{JianwenCVPR2018}, Pairwise~\cite{JohnsLD16}, GIFT~\cite{tmmbs2016}, PANORAMA~\cite{Sfikas17}, VRN~\cite{Brocknips2016}, RotationNet~\cite{AsakoCVPR2018}, PointNet++~\cite{nipspoint17}, T-L~\cite{Girdhar16}, LFD, Vconv-DAE~\cite{Sharma16}, 3DGAN~\cite{WuNIPS2016}, LGAN~\cite{PanosCVPR2018ICML}, and FNet~\cite{YaoqingCVPR2018}.

VIP-GAN significantly outperforms all its unsupervised competitors under ModelNet40, and some of them under ModelNet10, as shown by ``Our'', which is also the best result compared to eight top ranked supervised methods. For fair comparison, the result of VRN~\cite{Brocknips2016} is presented without ensemble learning, and the result of RotationNet\cite{AsakoCVPR2018} is presented with views taken by the default camera system orientation that is identical to the others. In addition, we try to train VIP-GAN under ShapeNet55 in unknown-test mode. Hence, we fix the parameters to extract features under ModelNet40 and ModelNet10, as shown by ``Our1(SN55)''. Although the results of LGAN\cite{PanosCVPR2018ICML} and FNet\cite{YaoqingCVPR2018} are better than ``Our1(SN55)'' under ModelNet10, it is inconclusive whether they are better than ours. This is because these methods are trained under a version of ShapeNet55 that contains more than 57,000 3D shapes, including a number of 3D point clouds. However, VIP-GAN is trained only under the 51,679 3D shapes from ShapeNet55 that are available for public download.

Finally, we explore whether ``Our'' could be further improved by more training shapes from ShapeNet55 in known-test mode, as shown by ``Our2(+SN55)''. However, with the existing parameters, only comparable results are obtained. Moreover, we evaluate VIP-GAN under ShapeNet55 in known-test mode using the same parameters with our best results under ModelNet10 in Table~\ref{table:parameters}, as shown in the rightmost column ``Our'' in Table~\ref{table:t10}. Similar to ``Our2(+SN55)'',  with the existing parameters, only comparable results are obtained by more training shapes from ModelNet40, as shown by ``Our+''.

\begin{table}
  \caption{The comparison of retrieval in terms of mAP under ModelNet40 and ModelNet10.}
  \label{table:retrieval}
  \centering
  \begin{tabular}{ccc}%llllllll
    \hline
    Methods & MN40 & MN10 \\
    \hline
    GeoImage & 51.30& 74.90  \\
    Pano & 76.81 & 84.18 \\
    MVCNN & 79.50 & -  \\
    GIFT & 81.94 & 91.12 \\
    RAMA & 83.45 & 87.39 \\
    Trip & 88.00 & - \\
    \hline
    Our & \textbf{89.23} & \textbf{90.69} \\
    Our1(SN55) & \textbf{87.66} & \textbf{90.09}\\
    Our2(+SN55) & \textbf{88.87} & \textbf{90.75} \\
    \hline
  \end{tabular}
\end{table}

\begin{figure}[!]
  \centering
  % the following command controls the width of the embedded PS file
  % (relative to the width of the current column)
  %\includegraphics[width=.95\linewidth, bb=39 696 126 756]{figures/definition3.eps}
   \includegraphics[width=\linewidth]{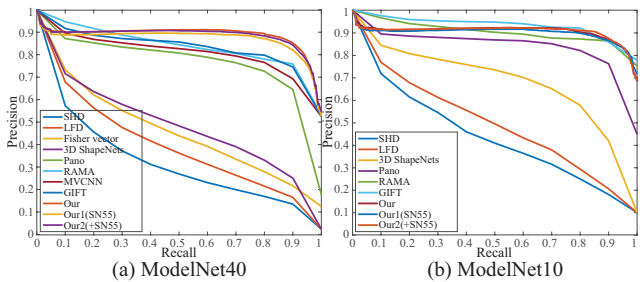}
  % replacing the above command with the one below will explicitly set
  % the bounding box of the PS figure to the rectangle (xl,yl),(xh,yh).
  % It will also prevent LaTeX from reading the PS file to determine
  % the bounding box (i.e., it will speed up the compilation process)
  % \includegraphics[width=.95\linewidth, bb=39 696 126 756]{sampleFig}
  %
  %
\caption{\label{fig:retrieval} The comparison of PR curves for retrieval under ModelNet40 and ModelNet10.}
\end{figure}

\noindent\textbf{Our novel implicit view aggregation. }The effect of our novel implicit view aggregation is first explored by visualization. In Fig.~\ref{fig:zeros}(a), we compare the training loss of our framework with a fixed, non-trainable $\bm{F}$ set to zero, and our trainable $\bm{F}$. Our approach is able to learn the characteristics of each shape to make up the missing information in each prediction, which reduces the training loss. The two losses show that the generator is getting to the Nash equilibrium.

%We visualize the effectiveness of our global feature implemented as a learned shape memory in Fig.~\ref{fig:zeros}(a) under ModelNet10.

In Fig.~\ref{fig:zeros}(b), we further evaluate the semantic meaning of our features by manipulating them algebraically, and visualizing the result via nearest neighbor retrieval in ModelNet10, as shown on the right. The retrieved shapes exhibit characteristics similar to both input shapes, such as the surface of the bed in the first row, and the bedhead in the second row.

%This shows that VIP-GAN can learn the semantic meaning of shapes.

\begin{table}
  \caption{The effects of our novel implicit view aggregation under ModelNet10.}
  \label{table:implicitcompare}
  \centering
  \begin{tabular}{c|c|c|c|c|c}%lllllllllllll
    \hline
       ACC&\multicolumn{2}{|c|}{Non-trainable $\bm{F}$}&\multicolumn{3}{c}{Trainable $\bm{F}$}\\
    \hline
     & MaxP&MeanP&MaxP&MeanP&Our\\
     \hline
     Ins&84.58&87.22&81.72&82.49&\textbf{94.05}\\
     Cla&83.95&87.38&80.60&81.73&\textbf{93.71}\\
    \hline
  \end{tabular}
\end{table}

%不训练全局特征，mean pooling的：testing instance acc: <0.872247>, testing class acc: <0.873837>
%不训练全局特征， max pooling的：testing instance acc: <0.845815>, testing class acc: <0.839488>
%训练全局特征，mean pooling的: testing instance acc: <0.824890>, testing class acc: <0.817302>
%训练全局特征，max pooling的: testing instance acc: <0.817181>, testing class acc: <0.805953>

Finally, we compare our implicit view aggregation with the widely used explicit view aggregation pooling under ModelNet10. Here, we use the output $\bm{h}_i$ of the encoder $\rm E$ as the feature of each view, and obtain the global feature of the shape by pooling all the $\bm{h}_i$ together with maxpooling and meanpooling, where each $\bm{h}_i$ is obtained with trainable $\bm{F}$ and non-trainable all zero $\bm{F}$. In Table~\ref{table:implicitcompare}, with trainable or non-trainable $\bm{F}$, our implicit view aggregation is always superior to the pooling. Without the support of trainable $\bm{F}$, the pooled features are pushed to be more discriminative than the ones with trainable $\bm{F}$ to minimize the loss, which makes the pooling results better. However, it is still not good enough to keep the loss as low as ours shown in Fig.~\ref{fig:zeros}(a).

\begin{table}[htb]
\centering
\caption{Retrieval and classification comparison in terms of Micro-averaged metrics under ShapeNetCore55.}  % ????????
    \begin{tabular}{c|c|c|c|c|c}  % ?????
     \hline
        & \multicolumn{5}{|c}{Micro} \\
     \hline
       Methods & P & R & F1 & mAP & NDCG  \\  % ?????п?
     \hline
       %\multirow{10}{*}{Tesing} & \multirow{10}{*}{} & \multicolumn{5}{|c|}{\multirow{10}{*}{}} & \multicolumn{5}{|c|}{\multirow{10}{*}{}}\\
       %\multirow{10}{*}{Tesing} & & & & & & & & & & & \\
       Kanezaki & 81.0 & 80.1 & \textbf{79.8} & 77.2 & 86.5 \\
       Zhou & 78.6 & 77.3 & 76.7 & 72.2 & 82.7 \\
       Tatsuma  & 76.5 & 80.3 & 77.2 & 74.9 & 82.8  \\
       Furuya  & \textbf{81.8} & 68.9 & 71.2 & 66.3 & 76.2  \\
       Thermos  & 74.3 & 67.7 & 69.2 & 62.2 & 73.2 \\
       Deng  & 41.8 & 71.7 & 47.9 & 54.0 & 65.4 \\
       Li  & 53.5 & 25.6 & 28.2 & 19.9 & 33.0 \\
       Mk  & 79.3 & 21.1 & 25.3 & 19.2 & 27.7 \\
       Su  & 77.0 & 77.0 & 76.4 & 73.5 & 81.5 \\
       Bai  & 70.6 & 69.5 & 68.9 & 64.0 & 76.5 \\
       Taco  & 70.1 & 71.1 & 69.9 & 67.6 & 75.6 \\
       Our  & 60.0 & \textbf{80.3} & 61.2 & \textbf{83.5} & \textbf{89.4} \\%@730
       Our+  & 60.0 & \textbf{80.3} & 61.2 & \textbf{83.6} & \textbf{89.5} \\%@731
        \hline
       \multicolumn{3}{c|}{Our accuracy} & \multicolumn{3}{|c}{\textbf{82.97}} \\
       \multicolumn{3}{c|}{Our+ accuracy} & \multicolumn{3}{|c}{\textbf{82.51}} \\
     \hline
   \end{tabular}
   \label{table:t10}
\end{table}

\noindent\textbf{Retrieval. }VIP-GAN is further evaluated in shape retrieval under ModelNet40, ModelNet10 and ShapeNet55, as shown in Table~\ref{table:retrieval}, Table~\ref{table:t10} and Table~\ref{table:t11}. The compared results include LFD, SHD, Fisher vector, 3D ShapeNets~\cite{Wu2015}, GeoImage~\cite{eccvSinha2017}, Pano~\cite{Bshi2015}, MVCNN~\cite{su15mvcnn}, GIFT~\cite{tmmbs2016}, RAMA~\cite{Sfikas17} and Trip~\cite{Xinweicvpr18}.

In these experiments, the 3D shapes in the test set are used as queries to retrieve the rest shapes in the same set, and mean Average Precision (mAP) is used as a metric. In addition, we employ global features involved in our classification results in Table~\ref{table:comparison} and Table~\ref{table:t10} for the retrieval experiments under the three benchmarks.

As shown in Table~\ref{table:retrieval}, our results of ``Our'' outperform all the compared results under ModelNet40, and sightly lower than the best results of $91.12$ by GIFT under ModelNet10. However, it is inconclusive whether GIFT outperforms VIP-GAN, since the dataset used by GIFT is formed by randomly selecting 100 shapes from each shape class, which is much simpler than the whole benchmark that we used. In addition, with trained by more shapes from ShapeNet55, the result of ``Our2'' under ModelNet10 is a little bit higher than the result of ``Our''. Their available PR curves under ModelNet40 and ModelNet10 are also compared in Fig.~\ref{fig:retrieval}.

In Table~\ref{table:t10} and Table~\ref{table:t11}, the results of ``Our'' outperform all the compared results under ShapeNet55. Besides Taco~\cite{s2018spherical} in Table~\ref{table:t10}, the compared results without reference are from SHREC2017 shape retrieval contest~\cite{3dor20171050} under ShapeNet55 with the same names, where micro-averaged and macro-averaged methods are employed to compute the metrics. Similar to ``Our2'' under ModelNet10, with trained by more shapes from ModelNet40, ``Our+'' is a little bit better than ``Our''.

\begin{table}[htb]
\centering
\caption{Retrieval comparison in terms of Macro-averaged metrics under ShapeNetCore55.}  % ????????
    \begin{tabular}{c|c|c|c|c|c}  % ?????
     \hline
        & \multicolumn{5}{|c}{Macro} \\
     \hline
       Methods & P & R & F1 & mAP & NDCG  \\  % ?????п?
     \hline
       %\multirow{10}{*}{Tesing} & \multirow{10}{*}{} & \multicolumn{5}{|c|}{\multirow{10}{*}{}} & \multicolumn{5}{|c|}{\multirow{10}{*}{}}\\
       %\multirow{10}{*}{Tesing} & & & & & & & & & & & \\
       Kanezaki & 60.2 & 63.9 & \textbf{59.0} & 58.3 & 65.6 \\
       Zhou & 59.2 & 65.4 & 58.1 & 57.5& 65.7\\
       Tatsuma  & 51.8 & 60.1 & 51.9 & 49.6 & 55.9 \\
       Furuya  & \textbf{61.8} & 53.3 & 50.5 & 47.7 & 56.3\\
       Thermos  & 52.3 & 49.4 & 48.4 & 41.8 & 50.2\\
       Deng  & 12.2 & 66.7 & 16.6 & 33.9 & 40.4\\
       Li  & 21.9 & 40.9 & 19.7 & 25.5 & 37.7\\
       Mk  & 59.8 & 28.3 & 25.8 & 23.2 & 33.7\\
       Su  & 57.1 & 62.5 & 57.5 & 56.6 & 64.0 \\
       Bai  & 44.4 & 53.1 & 45.4 & 44.7 & 54.8\\
       Our  & 18.9 & \textbf{81.2} & 24.0 & \textbf{69.2} & \textbf{83.7} \\%@730
       Our+  & 18.8 & \textbf{81.3} & 24.0 & \textbf{69.9} & \textbf{84.0} \\%@731
     \hline
   \end{tabular}
   \label{table:t11}
\end{table}

\section{Conclusions }
We proposed VIP-GAN, an approach for unsupervised 3D global feature learning by view inter-prediction that is capable of learning from fine-grained ``supervised'' information within the multi-view context of 3D shapes. Inspired by human perception of view-dependent patterns, VIP-GAN successfully learns more discriminative golbal features than state-of-the-art view-based methods that regard the multi-view context as a whole. With adversarial training, the global features can be learned more efficiently, which further improves their discriminability. In addition, our novel implicit aggregation enables VIP-GAN to learn within the multi-view context by effectively aggregating knowledge learned from multiple local view predictions across a view sequence. Our results show that VIP-GAN outperforms its unsupervised counterparts, as well as some top ranked supervised methods under large scale benchmarks in shape classification and retrieval.

\section{Acknowledgments}
Yu-Shen Liu is the corresponding author. This work was supported by National Key R\&D Program of China (2018YFB0505400), the National Natural Science Foundation of China (61472202), and Swiss National Science Foundation grant (169151). We thank all anonymous reviewers for their constructive comments.

\bibliographystyle{aaai}
\bibliography{../paper}

\end{document}